\documentclass{article} % For LaTeX2e
\usepackage{iclr2027_conference,times}

\usepackage{amsmath,amsfonts,bm,amssymb}

\def\eqref#1{equation~\ref{#1}}

\def\1{\bm{1}}

\DeclareMathAlphabet{\mathsfit}{\encodingdefault}{\sfdefault}{m}{sl}
\SetMathAlphabet{\mathsfit}{bold}{\encodingdefault}{\sfdefault}{bx}{n}

\usepackage{hyperref}
\usepackage{url}
\usepackage{graphicx}
\usepackage[table]{xcolor} % before tikz, which also loads xcolor
\usepackage{booktabs}
\usepackage{array}
\usepackage{tikz}
\usetikzlibrary{arrows.meta,calc}

\definecolor{baselinegray}{gray}{0.92}
\definecolor{losscolor}{HTML}{B2182B}
\newcommand{\sigpos}[1]{{\bfseries\boldmath #1}}                       % interval excludes zero, gain
\newcommand{\signeg}[1]{\textcolor{losscolor}{\bfseries\boldmath #1}} % interval excludes zero, loss
\newcommand{\ci}[2]{{\scriptsize\color{black!55}[#1,\,#2]}}
\title{Mining Agent Skills from Production Traces}

\author{Yue Ran Kang\textsuperscript{1,}\thanks{Equal contribution.}~, Colton Mikolajczyk\textsuperscript{1,}\rlap{$^*$}~, Chhaya Methani\textsuperscript{2,}\rlap{$^*$}~, Hazel Mak\textsuperscript{2,}\rlap{$^*$}~,\\
\textbf{Sahil Bhatnagar\textsuperscript{2}, Susheel Suresh\textsuperscript{2} \& Alejandro Gutierrez Munoz\textsuperscript{2}}\\
\textsuperscript{1}Massachusetts Institute of Technology \quad \textsuperscript{2}Microsoft Corporation\\
\texttt{\{kyueran,colto57\}@mit.edu} \quad \texttt{\{cmethani,hazelmak\}@microsoft.com}
}

\iclrfinalcopy

\begin{document}

\maketitle
\lhead{}

% main body

\begin{abstract}
    Agent skills that record procedural instructions are increasingly mined from execution traces rather than curated by hand. Skill-mining pipelines often use known task outcomes or feedback to guide skill construction. In production, reliable information on whether a run has succeeded may be unavailable. We study how the sampling of execution traces, access to success or failure information, and the form of the mined skills affect downstream task performance. Holding the mining pipeline fixed, we compare six combinations of mining evidence and skill forms. Mining evidence has three levels: successful trajectories only, successes and failures with their outcome labels, or the same mix with labels withheld. Skill form has two types: an ordered workflow plan, or a declarative ontology of entities, states, and policies. We evaluate the mined skills on two enterprise benchmarks, ThinkingBox-Bench and APEX-Agents. Analysis of task-level paired differences shows that the benefits of different configurations of mining evidence and skill forms depend on the enterprise domain. On ThinkingBox-Bench, paired differences show that workflows score better than ontology by $1.7$ pp, Goldilocks beats success-only evidence type by $2.4$ pp and Goldilocks blind simulating skills learnt without outcomes is worse by $3.1$ pp. APEX-Agents shows a moderate preference for ontologies and no clear preference between evidence regimes. Within each domain, task structure related constraints drive uneven performance with mined skills. These findings motivate tailoring meta-skills to the demands of the target tasks rather than adopting a one-size-fits-all approach.

\end{abstract}

\section{Introduction}
\label{sec:intro}
    In enterprise settings, agents often handle recurring business workflows such as returns, access requests, and account changes. Yet in a standard language model agent setup, a successful execution does not automatically become reusable guidance for later runs, making these recurring requests unnecessarily costly. For example, to resolve a customer’s return request, an agent may need to explore unfamiliar tools and records, infer how requests are routed, determine which eligibility rules apply, and discover the required state transitions through trial and error. When the session ends, what the agent learned about the environment and procedure is not preserved as guidance for the next run. The next return request therefore requires the agent to rediscover and execute the procedure again.

Skills address this problem by turning procedural knowledge from prior executions into reusable artifacts for future runs. A skill is a versioned set of instructions loaded into the agent’s context when relevant, without changing the model’s parameters \citep{anthropic2025skills}; this makes its influence inspectable, reversible, and suitable for enterprise deployment. A growing literature therefore studies how to mine skills directly from execution traces \citep{ni2026trace2skill,wang2024agentworkflowmemory}.

A skill is only as good as the quality of procedure it describes. A mined skill is a set of reusable instructions curated by a separate skill-mining agent that reviews prior task-agent execution traces. Its content therefore depends on which traces the miner sees and whether the miner knows which traces succeeded. Mined skills can improve performance, but they can also cause regressions when they overfit to the source traces by encoding unnecessary steps or overgeneralizing rules \citep{li2026skillsbench,dong2026agentskillsharmful}. In addition, many trace-mining methods select source traces based on whether prior runs succeeded or failed \citep{ni2026trace2skill,wang2026skillx}. In realistic production settings, however, success and failure outcomes may be unavailable because user traces cannot always be inspected or graded at scale for privacy and compliance reasons. This constraint motivates studying how mined skills change when the same traces are provided without outcome labels.

In this paper, we study how three things affect downstream task performance: which execution traces the miner is shown, what it is told about their outcomes, and what form the mined skill takes. We hold the mining pipeline fixed and compare six combinations of mining evidence and skill form. Mining evidence has three levels: \textbf{Success-only}, which shows the miner successful trajectories alone; \textbf{Goldilocks}, which shows successes and failures together with their outcome labels; and \textbf{Goldilocks-blind}, which shows that same mixture as goldilocks with the labels withheld. Goldilocks-blind represents the production-relevant constraint that outcome labels may be unavailable. Because it uses the same traces as Goldilocks but withholds their labels, the comparison isolates the contribution of label access rather than the inclusion of failure traces. Skill form has two types: a \textbf{workflow plan}, an ordered sequence of steps, and an \textbf{ontology}, a declarative account of the domain's entities, states, and policies that describes the world the agent acts in without dictating an order through it (see Figure ~\ref{fig:skill-forms}).

An additional constraint in production is the inability to tune the miner per domain or task family. Hence, we follow a minimal footprint approach to skill mining; the distillation instruction, or meta-skill, is authored once against a single agent's trajectories and then frozen. We apply this frozen meta-skill to traces from training tasks to produce skills, then evaluate those skills on test tasks (Figure~\ref{fig:method-pipeline}). We evaluate these mined skills on two enterprise benchmarks, ThinkingBox-Bench \citep{li2026successisntreliabilitythinkingbox} and APEX-Agents \citep{vidgen2026apexagents}. We run each task with and without the mined skill, and measure the paired difference in task success.

Our experiments show that skill representation and evidence selection have context-dependent effects on task success. Workflow representations yield positive point estimates relative to ontology representations in ThinkingBox-Bench in $4$ out of $5$ domains (Table \ref{tab:arm-comparisons-domains}). The aggregate difference is $1.7$ pp (Table \ref{tab:arm-comparisons}). Variance analysis showed that tasks with multi-entity interactions resolved better with ontology. Goldilocks outperforms success-only selection in ThinkingBox-Bench by $2.4$ pp suggesting that contrastive evidence may help resolve decision branches within tasks. Goldilocks-blind selection yields lower point estimates than Goldilocks across all reported settings (Table \ref{tab:arm-comparisons-domains}), with a matched regression of $3.1$ pp in ThinkingBox-Bench and $0.3$ pp in APEX-Agents ((Table \ref{tab:arm-comparisons})). 

Exploratory task-level analysis suggests that these aggregate results may mask differences in task requirements: ordered execution may favor workflow representations, while entity and policy-rich tasks may favor ontology representations. Similarly, comparing successful and failed runs may reveal missing decision points or completion conditions, but may help less when a task requires only one precise update and no additional changes. These patterns motivate task-sensitive skill design rather than a single universally preferred configuration. A practical takeaway for industry practitioners is to match skill design with task constraints. 

This paper makes three contributions.
\begin{enumerate}
\item \textbf{Controlled comparisons of skill form and mining evidence.} We compare workflow- and ontology-based skill mining while varying the evidence available during mining, measuring their effects on downstream enterprise task success.
\item \textbf {Quantifying the effect of outcome-label access.} Using matched trajectories for Goldilocks and outcome-blind generation, we isolate the effect of outcome-label visibility under a fixed experimental setup, examining a production-relevant information constraint. (\citep{zhang2026ace,xiong2025memorymanagement}). 
\item \textbf {Characterizing task-dependent gains and regressions.} We examine how outcomes vary across task families. The exploratory patterns suggest that effective skill design depends on task requirements, with no single configuration consistently performing best.
\end{enumerate}

\section{Related Work}
\label{sec:related}
    Prior work studies skill representation and generation evidence through controlled comparisons \citep{huang2026skillconsumption,jiang2026demystifying}, and evaluates downstream skill utility in enterprise settings \citep{kevin2026aces}. We situate our study along three axes: skill acquisition and evolution, skill representation, and downstream utility.

\paragraph{Skill acquisition and evolution.}
Skill acquisition methods differ in the evidence used to generate skills and whether skills are subsequently refined. Voyager learns executable skills, while ExpeL extracts natural-language insights \citep{wang2023voyager,zhao2024expel}; Agent Workflow Memory and Trace2Skill mine reusable routines and lessons from trajectories \citep{wang2024agentworkflowmemory,ni2026trace2skill}. SkillGen contrasts successes and failures, while other approaches use self-judged experiences or adapt without labeled supervision \citep{ma2026skillgen,ouyang2026reasoningbank,zhang2026ace}. Controlled studies examine evidence mixtures, generation methods, representations, and outcome labels \citep{huang2026skillconsumption,zhong2026skilllearnbench,jiang2026demystifying}. Beyond initial acquisition, ReMe and SkillEvolver update external knowledge from execution feedback \citep{cao2025reme,zhang2026skillevolver}; SkillRL couples library evolution with policy training \citep{xia2026skillrl}; and SkillOpt accepts skill edits when held-out validation improves \citep{yang2026skillopt}. Unlike these iterative approaches, we study one-shot skill generation using a frozen distillation instruction, without subsequent skill refinement or policy training. 

\paragraph{Skill form.} Prior work explores how reusable knowledge is expressed, from procedural guidance to structured representations. Memp compares stepwise and script-like guidance, while Agent Skill Induction compares programs and text under programmatic verification \citep{fang2025memp,wang2025asi}. \citet{feng2026skilltransfer} compare task- and subtask-level induction in text and code. 
Learning domain ontologies \citep{giglou2023llms4ol,lo2024ollm} is distinct from organizing skills in ontologies \citep{liang2026skillnet}. Our comparison instead concerns the content of the skill itself: an ordered workflow versus a declarative domain ontology that specifies entities, states, and policies without prescribing an execution sequence.

\paragraph{Downstream utility and task-dependent effects.} Skill availability, invocation, and artifact quality do not by themselves establish downstream benefit. Skill or Skip studies selective invocation, while More Skills, Worse Agents? examines selection errors and context effects in growing libraries \citep{chen2026skillorskip,song2026skillshadowing}. SkillFlow finds that frequent use need not help without suitable content \citep{li2026skillflow}. Downstream evaluations compare curated skills with no skills and distilled skills with raw traces under frozen deployment \citep{li2026skillsbench,lei2026skillevolbench}. ACES measures the added value of existing skills through paired enterprise trials under fixed execution and grading conditions \citep{kevin2026aces}. These evaluations also motivate looking beyond average gains: SWE-Skills-Bench reports limited average improvements and compatibility-related regressions, while other studies document uneven benefits and skill-induced failures \citep{han2026sweskillsbench,wang2026notallskills,dong2026agentskillsharmful}. The Regression Tax explicitly separates gains, regressions, and residual failures \citep{tank2026regressiontax}. We extend this evaluation perspective by examining task-family-level gains and regressions across skill representations and mining-evidence conditions, with exploratory analysis linking these differences to task requirements.

\section{Method}
\label{sec:method}
    \subsection{Setting and Design Overview}

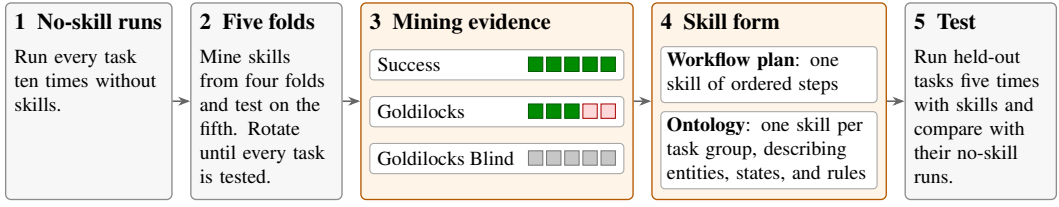
\begin{figure}[t]
  \centering
  \begin{tikzpicture}[
      font=\fontsize{7.5}{9}\selectfont,
      stage/.style={draw=black!45, line width=0.5pt, rounded corners=2pt, fill=black!3,
        anchor=north west, minimum height=2.6cm, inner sep=0pt},
      varied/.style={stage, draw=orange!70!black, fill=orange!9},
      head/.style={anchor=north west, font=\fontsize{8.5}{10}\selectfont\bfseries, inner sep=0pt},
      body/.style={anchor=north west, align=left, inner sep=0pt,
        execute at begin node={\hyphenpenalty=10000\exhyphenpenalty=10000}},
      chip/.style={draw=black!35, line width=0.4pt, rounded corners=1.5pt, fill=white,
        anchor=north west, inner sep=0pt},
      arrow/.style={-{Stealth[length=1.6mm]}, draw=black!60, line width=0.6pt},
      sq/.style={minimum size=1.9mm, inner sep=0pt, line width=0.35pt},
      pass/.style={sq, draw=green!35!black, fill=green!55!black},
      fail/.style={sq, draw=red!65!black, fill=red!15},
      hid/.style={sq, draw=black!55, fill=black!22},
    ]
    \node[stage, minimum width=2.2cm] (s1) at (0, 0) {};
    \node[stage, minimum width=2.0cm] (s2) at (2.45, 0) {};
    \node[varied, minimum width=3.6cm] (s3) at (4.7, 0) {};
    \node[varied, minimum width=3.1cm] (s4) at (8.55, 0) {};
    \node[stage, minimum width=2.0cm] (s5) at (11.9, 0) {};
    \foreach \a/\b in {s1/s2, s2/s3, s3/s4, s4/s5} \draw[arrow] (\a.east) -- (\b.west);

    \node[head] at (0.12, -0.14) {1\enspace No-skill runs};
    \node[body, text width=1.96cm] at (0.12, -0.62) {Run every task ten times without skills.};

    \node[head] at (2.57, -0.14) {2\enspace Five folds};
    \node[body, text width=1.76cm] at (2.57, -0.62) {Mine skills from four folds and test on the fifth. Rotate until every task is tested.};

    \node[head] at (4.82, -0.14) {3\enspace Mining evidence};
    \foreach \name/\y/\runs in {
        Success/-0.62/{pass,pass,pass,pass,pass},
        Goldilocks/-1.24/{pass,pass,pass,fail,fail},
        Goldilocks Blind/-1.86/{hid,hid,hid,hid,hid}} {
      \node[chip, minimum width=3.36cm, minimum height=0.4cm] at (4.82, \y) {};
      \node[anchor=west, inner sep=0pt] at (4.92, \y-0.2) {\name};
      \foreach \s [count=\i] in \runs
        \node[\s] at ({6.78 + 0.24*\i}, \y-0.2) {};
    }

    \node[head] at (8.67, -0.14) {4\enspace Skill form};
    \node[chip, minimum width=2.86cm, minimum height=0.74cm] at (8.67, -0.58) {};
    \node[body, text width=2.7cm] at (8.75, -0.65) {\textbf{Workflow plan}: one skill of ordered steps};
    \node[chip, minimum width=2.86cm, minimum height=1.02cm] at (8.67, -1.44) {};
    \node[body, text width=2.7cm] at (8.75, -1.51) {\textbf{Ontology}: one skill per task group, describing entities, states, and rules};

    \node[head] at (12.02, -0.14) {5\enspace Test};
    \node[body, text width=1.76cm] at (12.02, -0.62) {Run held-out tasks five times with skills and compare with their no-skill runs.};
  \end{tikzpicture}
  \caption{Study design. We vary two things, shaded: the evidence the miner sees and the form of the
    skill it writes, using the same mining prompts in every domain. Squares show the five runs taken
    from each selected training task: green passed, red failed, gray outcome hidden. Success runs come
    from tasks that pass 8--10 of their ten no-skill runs and Goldilocks runs from tasks that pass 3--7;
    Goldilocks Blind uses the same runs as Goldilocks with their outcomes hidden. The three evidence
    types and two skill forms give the six arms run in every domain.}
  \label{fig:method-pipeline}
\end{figure}

We study how the evidence a miner sees and the form of the skill it writes affect an agent's success on tasks the skill was not mined from. Each task has an initial environment state and a hidden binary grader, and every task belongs to one of the task groups defined before mining. A skill is a natural-language Markdown document with a routing description and operational guidance. A fixed meta-skill instructs the miner to derive such a skill from trajectory evidence. Holding the miner and the meta-skills fixed, we cross two skill forms with three levels of mining evidence, giving six arms. For every arm, we select evidence from no-skill runs of training tasks, mine skills once per fold, and test them on test tasks (Figure~\ref{fig:method-pipeline}; Section~\ref{sec:five-fold-evaluation}).

\subsection{Skill Forms and Mining Scope}

\paragraph{Workflow plans.} A workflow plan specifies task routing, identifier lookup, preconditions, policy checks, action order, completion checks, and recovery or clarification. Each workflow arm mines one global skill per fold from evidence that spans all training task groups.

\paragraph{Operational ontologies.} An operational ontology specifies entities, identifiers, relationships, states, valid transitions, completion signals, constraints, and the lookups needed before modifying records, and it guides the choice of actions without prescribing a sequence. Each ontology arm mines one skill from the training evidence of each task group, so comparing the two skill forms also changes the mining scope and the number of skills the agent can read.

Each form has one meta-skill, reused unchanged across domains, that treats traces as evidence rather than as instructions and includes no few-shot examples; test outcomes never guide the revision or selection of a skill (Appendices~\ref{app:skill-forms} and~\ref{app:mining-prompts}).

\subsection{Mining Evidence}

Repeated no-skill runs supply the mining evidence and also define task categories by baseline pass frequency. Each task has $B$ baseline runs with binary outcomes $y^{\mathrm{baseline}}_{x,r}$, and $p_x=\sum_{r=1}^{B}y^{\mathrm{baseline}}_{x,r}$ is its pass count. Let $c(x)$ denote the category assigned to task $x$, and let $\theta_{\mathrm{S}}$ and $\theta_{\mathrm{G}}$ denote the Success and Goldilocks pass-count cutoffs, where $0<\theta_{\mathrm{G}}<\theta_{\mathrm{S}}\leq B$. The three categories are
\begin{equation}
  \label{eq:baseline-categories}
  c(x)=
  \begin{cases}
    \text{Success}, & \theta_{\mathrm{S}} \leq p_x \leq B, \\
    \text{Goldilocks}, & \theta_{\mathrm{G}} \leq p_x < \theta_{\mathrm{S}}, \\
    \text{Failure}, & 0 \leq p_x < \theta_{\mathrm{G}}.
  \end{cases}
\end{equation}
Three levels of mining evidence control which tasks and runs the miner sees and whether it sees their outcomes:
\begin{itemize}
  \item \textbf{Success} (success-only): five labeled passing runs from each selected Success task.
  \item \textbf{Goldilocks}: three passing and two failing runs, with labels, from each selected Goldilocks task.
  \item \textbf{Goldilocks Blind} (outcome-blind): exactly the same tasks, runs, and order as Goldilocks, without labels.
\end{itemize}

These levels separate outcome-label access from evidence selection. Goldilocks and Goldilocks Blind mine the same runs and differ only in the outcome labels and in the mining instructions that go with them (Appendix~\ref{app:mining-prompts}), so their comparison measures the value of outcome labels, together with the instructions that use them, on fixed evidence. Success and Goldilocks differ in both the task category and the mix of passing and failing runs, so their comparison covers two complete selection strategies. Goldilocks Blind withholds labels, as a deployed agent must, but its tasks and runs are still selected using outcomes; Appendix~\ref{app:random-blind} evaluates a fourth level that removes this selection. The miner never sees hidden grading material (Appendix~\ref{app:mining-protocol}).

\section{Experimental Setup}
\label{sec:setup}
    \subsection{Benchmarks}

ThinkingBox-Bench contains 516 enterprise-support tasks in five domains: neobank IT support (104 tasks), consulting helpdesk (107), car insurance (101), retail e-commerce (100, of which we evaluate 98), and hotel booking management (104) \citep{li2026successisntreliabilitythinkingbox}. Agents resolve customer or employee requests with ticketing and domain tools, and a simulated user supplies private information. A run passes only if its final backend state matches a golden state, so a correct decision recorded with nonstandard field values fails. Each domain's two scenario groups, or three in consulting, define its ontology scopes (Table~\ref{tab:thinkingbox-datasets}).

APEX-Agents contains 480 professional-services tasks, 160 each in investment banking, law, and management consulting \citep{vidgen2026apexagents}. From a request and seeded workspace files, the agent produces a deliverable, such as a financial analysis or legal memo, in an isolated workspace with the benchmark's nine Model Context Protocol servers. A run passes only if the strict binary grader accepts the deliverable against private rubric criteria, and the three domains are the ontology scopes.

\subsection{Agent and Miner}

Both benchmarks run the agent in the GitHub Copilot runtime (Copilot CLI 1.0.79-8 for ThinkingBox-Bench) with GPT-5.4 at medium reasoning effort, and the same model mines every skill in one miner call; all runs use seed 42. ThinkingBox-Bench's simulated users and rubric judge use GPT-5. ThinkingBox-Bench runs time out after 3,600 seconds, with at most 40 tool calls and 900 seconds per turn, and APEX-Agents runs have two million tokens and 1,800 seconds (Appendix~\ref{app:harness-details}).

\subsection{Mining Configuration}

Each task has $B=10$ earlier no-skill baseline runs, and the cutoffs $\theta_{\mathrm{S}}=8$ and $\theta_{\mathrm{G}}=3$ categorize tasks with 8--10, 3--7, and 0--2 passes as Success, Goldilocks, and Failure tasks, respectively (Equation~\ref{eq:baseline-categories}). All arms share five folds, stratified by task scenario and baseline category unless a frozen assignment is reused; each fold's skills are mined from the other four folds, so every task is tested once in each arm. Each skill draws its evidence from up to ten eligible training tasks of its scope and five runs of each, at most 50 trajectories, selected by seeded hash rankings (Appendix~\ref{app:mining-protocol}). In each fold, a workflow arm mines one global skill and an ontology arm mines one skill per task group. Across the five folds and three evidence levels, this produces 45 skills in each of the four ThinkingBox-Bench domains with two scenario groups, and 60 in consulting and in APEX-Agents, which have three task groups each.

\subsection{Evaluation Protocol and Metrics}
\label{sec:five-fold-evaluation}

Each test run receives only its arm's fold-specific skills, either the workflow skill or all task-group ontologies, installed as native Copilot skills: the agent sees each skill's name and routing description and chooses which to read, a read is recorded only when the runtime opens a skill, and skill content is never inserted into the task prompt. Every test task runs five times per arm, giving 15{,}420 skill-enabled ThinkingBox-Bench runs and 14{,}400 APEX-Agents runs across the six arms. The controls are the ten earlier no-skill runs in neobank, consulting, car insurance, and APEX-Agents, which also supply the mining evidence, and five no-skill runs collected for this study in retail and hotel booking. Controls share every setting of the skill runs, although earlier controls cannot rule out drift in the hosted model (Appendix~\ref{app:harness-details}).

The metric is the binary pass rate. Let $K_{\mathrm{T}}$ and $K_{\mathrm{B}}$ denote the numbers of treatment and baseline runs per task. Let $y^{\mathrm{T}}_{x,r}$ and $y^{\mathrm{B}}_{x,r}$ be the outcomes of task $x$ in repetition $r$ of the treatment runs, with the skill, and of the baseline runs, without it. For a set $\mathcal{D}$ of $N$ test tasks, the per-task difference and the lift are
\begin{equation}
	\label{eq:paired-lift}
	d_x=\frac{1}{K_{\mathrm{T}}}\sum_{r=1}^{K_{\mathrm{T}}}y^{\mathrm{T}}_{x,r}
	-\frac{1}{K_{\mathrm{B}}}\sum_{r=1}^{K_{\mathrm{B}}}y^{\mathrm{B}}_{x,r},
	\qquad
	\Delta=\frac{1}{N}\sum_{x\in\mathcal{D}}d_x.
\end{equation}
Pairing is by task rather than by repetition, every task is weighted equally, and comparisons between two arms use the same per-task differences with the second arm in place of the control. Runs that end at a limit or in an agent error count as ordinary attempts, as do runs without skill reads, whereas system errors are rejected. We report lifts and arm differences in percentage points with 95\% intervals from bootstraps that resample whole tasks and keep all arms of a task together. Benchmark aggregates weight domains equally and are not pooled across benchmarks, and ThinkingBox-Bench's aggregate lifts combine domain intervals with a normal approximation. Appendices~\ref{app:detailed-results}--\ref{app:pipeline-observability} add Holm-adjusted significance tests, task-level analyses, and a pipeline audit. The Random Blind extension (Appendix~\ref{app:random-blind}) keeps this protocol except for its evidence selection.

\section{Results}
\label{sec:results}
    We evaluate six combinations of two skill forms (workflow plans and ontologies) and three mining evidence regimes (Success, Goldilocks, and Goldilocks Blind) on tasks not used for skill mining, with five runs per task and arm. We first compare all six skill-enabled arms with the no-skill baseline, then compare skill forms and evidence choices directly. Appendix~\ref{app:random-blind} reports the Random Blind extension, which simulates skill mining in a production setting by randomly sampling training tasks and omitting outcome labels.

\subsection{Core Outcomes: Gains Over Baseline}
\label{sec:results-overall}

We test whether mined skills improve average pass rate over the no-skill baseline. Workflow plans show the clearest gains on ThinkingBox-Bench, whose mostly operational tasks may benefit from explicit action sequences. Ontologies show the clearest gains on APEX-Agents, whose expert-curated, knowledge-intensive tasks may benefit from organized domain concepts, relationships, and rules.

\begin{table}[h]
  \caption{Aggregate lift of each skill-mining arm over the no-skill baseline. Domains receive equal weight within each benchmark. The shaded row gives the baseline pass rate (\%); other entries are paired differences in percentage points (Equation~\ref{eq:paired-lift}) with 95\% intervals beneath. Bold: the interval excludes zero (unadjusted). ThinkingBox-Bench intervals use a normal approximation combining domain uncertainty and treating domains as independent; APEX-Agents intervals use task bootstraps within domains. Five runs are evaluated per task and arm; APEX-Agents controls use ten historical runs. APEX-Agents grades precede judge-service corrections (Appendix~\ref{app:detailed-results}).}
  \label{tab:uplift-baseline}
  \centering
  \small
  \setlength{\tabcolsep}{8pt}
  \begin{tabular}{l cc}
    \toprule
    Benchmark & ThinkingBox-Bench & APEX-Agents \\
    & {\footnotesize(5 domains)} & {\footnotesize(3 domains)} \\
    \midrule
    \rowcolor{baselinegray} No skill: pass rate (\%) & 48.9 & 26.6 \\
    \midrule
    \multicolumn{3}{l}{\textit{Workflow plan: lift over baseline}} \\
    \quad Success & \sigpos{+2.4} & +1.1 \\
    & \ci{0.2}{4.5} & \ci{$-0.3$}{2.5} \\[1pt]
    \quad Goldilocks & \sigpos{+3.8} & +0.4 \\
    & \ci{1.5}{6.1} & \ci{$-1.0$}{1.8} \\[1pt]
    \quad Goldilocks Blind & +1.0 & +0.4 \\
    & \ci{$-1.2$}{3.1} & \ci{$-0.8$}{1.6} \\[1pt]
    \midrule
    \multicolumn{3}{l}{\textit{Operational ontology: lift over baseline}} \\
    \quad Success & $-0.5$ & \sigpos{+1.6} \\
    & \ci{$-2.6$}{1.5} & \ci{0.2}{3.0} \\[1pt]
    \quad Goldilocks & \sigpos{+2.9} & \sigpos{+1.5} \\
    & \ci{0.6}{5.2} & \ci{0.1}{3.0} \\[1pt]
    \quad Goldilocks Blind & $-0.4$ & +0.9 \\
    & \ci{$-2.4$}{1.5} & \ci{$-0.6$}{2.4} \\[1pt]
    \bottomrule
  \end{tabular}
\end{table}

On ThinkingBox-Bench, workflow Goldilocks shows the largest estimated gain against the no-skill baseline, while workflow Success and ontology Goldilocks also show statistically significant improvements. Our trajectory analysis suggests that explicit action sequences encoded in workflow skills help agents complete linked record changes rather than stop at decision boundaries. Ontologies, on the other hand, may help preserve constraints across records, but their aggregate benefit is less consistent across evidence regimes.

On APEX-Agents, ontologies show gains with both, Success and Goldilocks evidence, while workflow Success has a slightly smaller, more uncertain estimated benefit. Inspected ontology wins involve selecting authoritative sources or distinguishing independent legal requirements, suggesting a benefit from organizing domain knowledge.

Goldilocks Blind shows no clear aggregate improvement against no-skill baseline across either benchmark or skill form. This is consistent with labels helping the miner distinguish useful guidance from unsuccessful behavior.

We also investigate domain-level performance in Table~\ref{tab:uplift-baseline-domains} that show uneven gains from skills across domains. $4$ domains in ThinkingBox-Bench show statistically significant gains with atleast one of the skill configurations, and $1$ shows no statistically significant improvement with skills (i.e. Hotel). On the APEX-Agents, only $1$ domain shows a positive statistically significant improvement (see Management Consulting), the other $2$ show moderate gains. This confirms the inconsistent skill related gains reported in Related Work in Section \ref{sec:related}. 

\begin{table}[h]
  \caption{Lift over the no-skill baseline within each ThinkingBox-Bench and APEX-Agents domain. Shaded rows give baseline pass rates (\%); other entries give paired lift in percentage points with 95\% task-bootstrap intervals beneath. Bold: the interval excludes zero (unadjusted); red: a loss. Each test task has five runs per arm; APEX-Agents controls use ten historical runs. Goldilocks Blind uses the Goldilocks traces with outcome labels removed. APEX-Agents grades precede judge-service corrections. Pass counts and inference details appear in Appendix~\ref{app:detailed-results}.}
  \label{tab:uplift-baseline-domains}
  \centering
  \footnotesize
  \setlength{\tabcolsep}{1.5pt}
  \resizebox{\linewidth}{!}{%
  \begin{tabular}{@{}l*{8}{c}@{}}
    \toprule
    & \multicolumn{5}{c}{ThinkingBox-Bench} & \multicolumn{3}{c}{APEX-Agents} \\
    \cmidrule(lr){2-6}\cmidrule(lr){7-9}
    Split & Neobank & Consulting & \shortstack{Car\\insurance} & Retail & Hotel & \shortstack{Investment\\banking} & Law & \shortstack{Management\\consulting} \\
    \midrule
    \rowcolor{baselinegray} \shortstack[l]{No skill:\\pass rate (\%)} & 45.0 & 47.5 & 57.1 & 51.0 & 44.0 & 32.6 & 18.6 & 28.5 \\
    \midrule
    \multicolumn{9}{l}{\textit{Workflow plan}} \\
    \quad Success & +5.4 & \sigpos{+4.9} & +0.7 & +1.0 & $-0.2$ & \signeg{$-2.2$} & +1.4 & \sigpos{+4.1} \\
    & \ci{$-1.0$}{11.8} & \ci{1.2}{8.7} & \ci{$-3.7$}{5.0} & \ci{$-4.1$}{6.3} & \ci{$-4.4$}{3.8} & \ci{$-4.3$}{$-0.1$} & \ci{$-0.8$}{3.5} & \ci{1.2}{7.1} \\[1pt]
    \quad Goldilocks & \sigpos{+7.3} & +0.7 & +3.7 & \sigpos{+7.1} & +0.2 & $-0.7$ & $-0.6$ & +2.5 \\
    & \ci{1.4}{13.4} & \ci{$-3.3$}{4.8} & \ci{$-1.4$}{8.7} & \ci{2.0}{12.7} & \ci{$-4.8$}{5.4} & \ci{$-3.4$}{2.0} & \ci{$-2.8$}{1.4} & \ci{$-0.1$}{5.1} \\[1pt]
    \shortstack[l]{\quad Goldilocks\\\quad Blind} & +4.8 & \signeg{$-3.6$} & \sigpos{+7.0} & $-0.6$ & $-2.9$ & $-1.1$ & +1.3 & +1.0 \\
    & \ci{$-0.8$}{10.6} & \ci{$-6.7$}{$-0.4$} & \ci{2.2}{11.9} & \ci{$-6.5$}{5.1} & \ci{$-7.3$}{1.3} & \ci{$-3.1$}{0.9} & \ci{$-0.4$}{3.1} & \ci{$-1.4$}{3.4} \\[1pt]
    \midrule
    \multicolumn{9}{l}{\textit{Operational ontology}} \\
    \quad Success & +3.7 & $-0.2$ & $-0.3$ & $-1.4$ & \signeg{$-4.4$} & +1.4 & +1.3 & +2.1 \\
    & \ci{$-0.7$}{8.1} & \ci{$-5.0$}{4.7} & \ci{$-5.0$}{4.5} & \ci{$-6.1$}{3.3} & \ci{$-8.7$}{$-0.4$} & \ci{$-1.2$}{4.1} & \ci{$-0.6$}{3.1} & \ci{$-0.5$}{4.8} \\[1pt]
    \quad Goldilocks & +4.8 & +0.6 & +2.7 & +1.4 & +5.0 & $-0.1$ & +2.1 & +2.5 \\
    & \ci{$-0.6$}{10.4} & \ci{$-3.9$}{5.0} & \ci{$-2.4$}{7.6} & \ci{$-3.9$}{6.7} & \ci{$-0.4$}{10.8} & \ci{$-2.6$}{2.4} & \ci{$-0.1$}{4.4} & \ci{$-0.1$}{5.1} \\[1pt]
    \shortstack[l]{\quad Goldilocks\\\quad Blind} & +2.7 & $-0.7$ & $-1.5$ & $-0.4$ & $-2.1$ & $-2.2$ & +1.6 & \sigpos{+3.3} \\
    & \ci{$-2.4$}{7.9} & \ci{$-4.5$}{3.1} & \ci{$-5.7$}{2.8} & \ci{$-4.5$}{3.7} & \ci{$-6.5$}{2.1} & \ci{$-4.7$}{0.3} & \ci{$-0.7$}{4.0} & \ci{0.3}{6.3} \\
    \bottomrule
  \end{tabular}
  }
\end{table}

Car insurance is one of the only domains to show its largest estimated gain with workflow Goldilocks Blind. Insurance decisions depend on the customer's policy and circumstances, so rules learnt from a few examples may not generalize. In inspected trajectories, the labeled skill overgeneralized a state-specific requirements, suggesting that avoiding such overgeneralization may explain some blind-arm gains.

APEX-Agents has no consistent leading arm; ontology Success has moderate positive lift across all $3$ domains. In management consulting, all configurations provide uplift, with workflow Success and ontology Goldilocks Blind showing comparable gains that are statistically significant (unadjusted). The latter provides the only statistically significant Blind gain among the APEX-Agents domains. In an inspected revenue-loss task, these results suggest that, in some settings, skills mined without outcome labels can offer benefits alongside those mined from successful trajectories.

Across both benchmarks, we see some configurations with statistically significant negative impact of skills across domains (see Consulting, Hotel and Investment banking in Table \ref{tab:uplift-baseline-domains}). This is in alignment with a growing body of research finding similar results across other domains like coding \cite{dong2026agentskillsharmful}.

\subsection{Comparisons Between Arms}
\label{sec:results-comparisons}

We compare arms directly to ask which skill forms and evidence choices work better. 

\begin{table}[h]
  \caption{Aggregate paired comparisons between skill forms and evidence conditions. Entries are first-minus-second pass-rate differences in percentage points with 95\% task-bootstrap intervals beneath. Skill-form contrasts average over three evidence conditions; evidence contrasts average over two skill forms. Domains receive equal weight within each benchmark. Negative values favor ontologies in the first row and Goldilocks in the second; positive values favor labeled Goldilocks in the third. Bold: the interval excludes zero (unadjusted). Each task has five runs per arm. APEX-Agents grades precede judge-service corrections (Appendix~\ref{app:detailed-results}).}
  \label{tab:arm-comparisons}
  \centering
  \small
  \setlength{\tabcolsep}{8pt}
  \begin{tabular}{l cc}
    \toprule
    Benchmark & ThinkingBox-Bench & APEX-Agents \\
    & {\footnotesize(5 domains)} & {\footnotesize(3 domains)} \\
    \midrule
    Workflow plan $-$ ontology & \sigpos{+1.7} & $-0.7$ \\
    & \ci{0.3}{3.1} & \ci{$-1.6$}{0.2} \\[1pt]
    Success $-$ Goldilocks & \sigpos{$-2.4$} & +0.4 \\
    & \ci{$-4.1$}{$-0.8$} & \ci{$-0.7$}{1.5} \\[1pt]
    Goldilocks $-$ Goldilocks Blind & \sigpos{+3.1} & +0.3 \\
    & \ci{1.4}{4.7} & \ci{$-0.8$}{1.4} \\
    \bottomrule
  \end{tabular}
\end{table}

\paragraph{Skill form.} Workflow plans have an average advantage on ThinkingBox-Bench, while APEX-Agents leans toward ontologies with less certainty. Workflow plans may help agents complete all required steps instead of stopping partway through, while ontologies may help them identify which rules apply. The skill-form comparison averages results across the three evidence-selection arms for each representation. The workflow and ontology configurations also differ in task grouping, skill count, and evidence quantity, so this comparison does not isolate representation alone (Section~\ref{sec:method}).

\paragraph{Evidence selection.} Goldilocks outperforms Success on ThinkingBox-Bench, whereas APEX-Agents shows no clear preference between them. The evidence-selection comparison averages results across tasks and the two skill-form arms for each evidence type. Failed examples may reveal skipped steps, but additional guidance can also prompt actions the task does not require. The observed difference reflects the evidence-selection strategy as a whole; the contribution of failed traces cannot be separated from changes in task selection and trace composition.

\paragraph{Outcome labels.} Labeled Goldilocks outperforms Goldilocks Blind on ThinkingBox-Bench, but not clearly on APEX-Agents. The comparison averages results across tasks and the two skill-form arms for each label condition. Labels may help miners identify required approvals and payment rules; \citet{jiang2026demystifying} likewise find a label benefit for pools of three successes and two failures. The observed difference reflects the labeled versus blind mining conditions as a whole: traces are held fixed, but label access and mining instructions change together, so this comparison does not isolate label access alone.

\begin{table}[h]
  \caption{Paired arm comparisons within each ThinkingBox-Bench and APEX-Agents domain. Entries are first-minus-second pass-rate differences on the same test tasks, in percentage points, with 95\% task-bootstrap intervals beneath. Skill-form contrasts average over three evidence conditions; evidence contrasts average over two skill forms. Negative values favor ontologies in the first row and Goldilocks in the second; positive values favor labeled Goldilocks in the third. Bold: the interval excludes zero (unadjusted). Each task has five runs per arm. APEX-Agents grades precede judge-service corrections; aggregate contrasts appear in Table~\ref{tab:arm-comparisons}.}
  \label{tab:arm-comparisons-domains}
  \centering
  \footnotesize
  \setlength{\tabcolsep}{1.5pt}
  \resizebox{\linewidth}{!}{%
  \begin{tabular}{@{}l*{8}{c}@{}}
    \toprule
    & \multicolumn{5}{c}{ThinkingBox-Bench} & \multicolumn{3}{c}{APEX-Agents} \\
    \cmidrule(lr){2-6}\cmidrule(lr){7-9}
    Split & Neobank & Consulting & \shortstack{Car\\insurance} & Retail & Hotel & \shortstack{Investment\\banking} & Law & \shortstack{Management\\consulting} \\
    \midrule
    Workflow plan & +2.1 & +0.8 & \sigpos{+3.5} & +2.7 & $-0.4$ & $-1.0$ & $-1.0$ & $-0.1$ \\
    \quad $-$ ontology & \ci{$-1.2$}{5.4} & \ci{$-2.4$}{4.0} & \ci{0.4}{6.7} & \ci{$-0.2$}{5.6} & \ci{$-3.5$}{2.6} & \ci{$-2.5$}{0.5} & \ci{$-2.5$}{0.4} & \ci{$-1.8$}{1.7} \\[1pt]
    Success & $-1.5$ & +1.7 & $-3.0$ & \sigpos{$-4.5$} & \sigpos{$-4.9$} & 0.0 & +0.6 & +0.6 \\
    \quad $-$ Goldilocks & \ci{$-5.3$}{2.1} & \ci{$-1.7$}{5.0} & \ci{$-7.2$}{1.3} & \ci{$-7.7$}{$-1.4$} & \ci{$-9.0$}{$-1.0$} & \ci{$-1.7$}{1.8} & \ci{$-1.2$}{2.3} & \ci{$-1.5$}{2.7} \\[1pt]
    Goldilocks & +2.3 & +2.8 & +0.4 & \sigpos{+4.8} & \sigpos{+5.1} & +1.3 & $-0.7$ & +0.4 \\
    \shortstack[l]{\quad $-$ Goldilocks\\\quad\phantom{$-$ }Blind} & \ci{$-1.3$}{5.9} & \ci{$-0.7$}{6.4} & \ci{$-3.5$}{4.2} & \ci{0.8}{8.9} & \ci{1.9}{8.5} & \ci{$-0.7$}{3.2} & \ci{$-2.3$}{0.9} & \ci{$-1.8$}{2.5} \\
    \bottomrule
  \end{tabular}
  }
\end{table}

\textbf{Domain-level comparisons} show statistically significant differences between arms in $3$ of $5$ ThinkingBox-Bench domains and none of the $3$ APEX-Agents domains (Table~\ref{tab:arm-comparisons-domains}). Car insurance favors workflow plans over ontologies, while both retail and hotel favor Goldilocks over Success and labeled Goldilocks over Goldilocks Blind. Differences in the remaining domains are less certain.

Our findings show that no single skill form or evidence regime consistently performs best across both benchmarks. The practical lesson is to evaluate alternative meta-skill configurations on representative tasks from the target deployment, rather than assume one configuration will work best everywhere. Because the illustrative cases were selected after outcomes were known, their suggested explanations should be treated as hypotheses. Validate configuration choices on held-out tasks, weighing the magnitude and uncertainty of performance gains alongside whether agents complete required steps and apply the correct rules (Appendix~\ref{app:task-level}).

\section{Conclusion}
\label{sec:conclusion}
    We studied how mining evidence, outcome-label access, and skill representation affect enterprise-agent performance under a fixed mining pipeline. On ThinkingBox-Bench, workflow plans and labeled Goldilocks evidence performed better on average, while APEX-Agents showed smaller, less conclusive differences, with no configuration consistently performing best across both benchmarks (Section 5.2). Task-level analysis suggests that gains and regressions may depend on required actions, entity relationships, and policy constraints, but does not establish a rule for choosing skills on unseen tasks (Examples shared in Appendix A.1, Figure~\ref{fig:skill-forms}). These findings motivate task-sensitive skill mining and validation on held-out tasks, rather than assuming that reusable guidance necessarily improves execution. Future work will investigate task-level skill evolution and adaptive selection from pools of workflow and ontology skills to better match task requirements.

% acknowledgements

\subsubsection*{Acknowledgments}

This work was sponsored by Microsoft and conducted as part of the Analytics Capstone of the MIT Master of Business Analytics program, which the MIT Sloan School of Management established with the MIT Operations Research Center. We thank the Microsoft Copilot Studio team for their thoughtful feedback and constructive discussions throughout the research. We are grateful to Soufiane Loukili, who leads the Copilot Studio team, for technical guidance and project sponsorship. We also thank the MIT faculty advisor, James Butler, for guidance and feedback throughout the project, and William McEntee, who leads the Analytics Capstone, for making this collaboration possible.

% references

\clearpage
\bibliography{iclr2027_conference}
\bibliographystyle{iclr2027_conference}

% appendix

\appendix
\section{Appendix}
\label{sec:appendix}
    \subsection{What Workflow Plans and Ontologies Look Like}
\label{app:skill-forms}

\begin{figure}[h]
  \centering
  \begin{tikzpicture}[
      font=\fontsize{7.5}{9}\selectfont,
      panel/.style={anchor=north west, font=\fontsize{8.5}{10}\selectfont, inner sep=0pt},
      wstep/.style={draw=black!45, line width=0.5pt, rounded corners=2pt, fill=black!4, minimum height=0.62cm,
        minimum width=2.5cm, anchor=north west, align=center, inner sep=2pt},
      wstop/.style={wstep, fill=black!72, draw=black!72, text=white, font=\fontsize{7.5}{9}\selectfont\bfseries},
      box/.style={draw=black!30, line width=0.45pt, rounded corners=2pt, fill=white, anchor=north west,
        align=left, inner sep=4pt,
        execute at begin node={\hyphenpenalty=10000\exhyphenpenalty=10000}},
      entity/.style={draw=black!55, line width=0.6pt, rounded corners=10pt, fill=black!4, align=center,
        inner sep=5pt, minimum width=2.0cm},
      good/.style={box, draw=green!45!black, line width=0.7pt, fill=green!4},
      bad/.style={box, draw=red!60!black, line width=0.7pt, fill=red!4},
      arrow/.style={-{Stealth[length=1.6mm]}, draw=black!60, line width=0.6pt},
      pill/.style={rounded corners=4pt, inner xsep=4pt, inner ysep=1.5pt, text=white,
        font=\fontsize{6.5}{7.5}\selectfont\bfseries},
    ]
    \node[panel] (pa) at (0, 0) {\textbf{(a) Workflow plan:} ordered lifecycle steps and a stopping point};
    \foreach \i/\t in {0/1\enspace Discover facts, 1/2\enspace Create ticket, 2/3\enspace Perform action, 3/4\enspace Update exact state}
      \node[wstep] (w\i) at ({2.85*\i}, -0.45) {\t};
    \node[wstop] (w4) at (11.4, -0.45) {5\enspace Stop};
    \foreach \a/\b in {w0/w1, w1/w2, w2/w3, w3/w4} \draw[arrow] (\a.east) -- (\b.west);

    \node[box, text width=6.4cm] (wex) at ($(w0.south west)+(0,-0.3)$) {\textbf{Example request.}
      \emph{``My laptop shut off after I spilled coffee on it and won't turn back on.''}\\[3pt]
      \textbf{Prescribed path:} open a ticket $\rightarrow$ request manager approval $\rightarrow$ record the approval
      and approver $\rightarrow$ set the request to pending $\rightarrow$ stop};
    \node[box, text width=6.42cm] (wmine) at (7.2, 0 |- wex.north) {\textbf{Mined workflow excerpt} (neobank):\\[3pt]
      \emph{``For approval-gated work, record the case, create the approval request, persist approver and expiry
      details, and stop at a pending state.''}};

    \node[panel] (pb) at ($(wex.south west)+(0,-0.5)$) {\textbf{(b) Operational ontology:} entities, relationships,
      and the changes they allow};

    \node[box, text width=6.42cm] (oex) at ($(pb.south west)+(7.2,-0.3)$) {\textbf{Example request.}
      \emph{``How do I set up VPN to access company systems?''}\\[3pt]
      The ontology relates the requester's role and current access to the resource's policy; for a standard
      employee, the valid response is setup guidance in an informational ticket, not an access change.};
    \node[box, text width=6.42cm] (omine) at ($(oex.south west)+(0,-0.25)$) {\textbf{Mined ontology excerpts}
      (car insurance):\\[3pt]
      \emph{``Entities: requester, customer record, policy, policy role, insured vehicle, insured driver, document
      request, secure document link, claim intake, claim.''}\\[3pt]
      \emph{``Authorized relationship: the requester must map cleanly to the policy context for the action being
      taken.''}};

    \node[good, text width=3.15cm] (resA) at ($(pb.south west)+(3.55,-0.4)$) {\textbf{Resource A}\\
      current access: none\\requested access: write\\policy: may grant\\[2pt]\textcolor{green!35!black}{\textbf{Valid}}
      $\rightarrow$ change state};
    \node[bad, text width=3.15cm] (resB) at ($(resA.south west)+(0,-0.5)$) {\textbf{Resource B}\\
      current access: read\\requested access: write\\policy: forbids write\\[2pt]\textcolor{red!55!black}{\textbf{Invalid}}
      $\rightarrow$ reject, keep read};
    \coordinate (mid) at ($(resA.south west)!0.5!(resB.north west)$);
    \node[entity] (person) at (1.05, 0 |- mid) {\textbf{Requester}\\[1pt]role and\\current access};
    \draw[arrow, draw=green!45!black, line width=0.9pt] (person.east) -- (resA.west);
    \draw[arrow, draw=red!60!black, line width=0.9pt, dash pattern=on 3pt off 2pt] (person.east) -- (resB.west);
    \node[pill, fill=green!45!black] at ($(resA.west)+(-0.9,0.42)$) {feasible};
    \node[pill, fill=red!60!black] at ($(resB.west)+(-0.9,-0.42)$) {infeasible};
  \end{tikzpicture}
  \caption{What the two skill forms encode, drawn schematically; quoted excerpts are verbatim from skills mined in
    this study. (a) A workflow plan is procedural: it orders the steps of a request's lifecycle, from discovering
    facts to updating the final state, and says when to stop. (b) An operational ontology is declarative: it names
    the entities a request involves, such as the requester and the resources they ask about, the relationships among
    them, such as current access, requested access, and the governing policy, and the state changes those
    relationships permit. The same requester can validly gain access to one resource but must be refused, with
    existing access preserved, for another. The ontology leaves the order of tool calls to the agent.}
  \label{fig:skill-forms}
\end{figure}
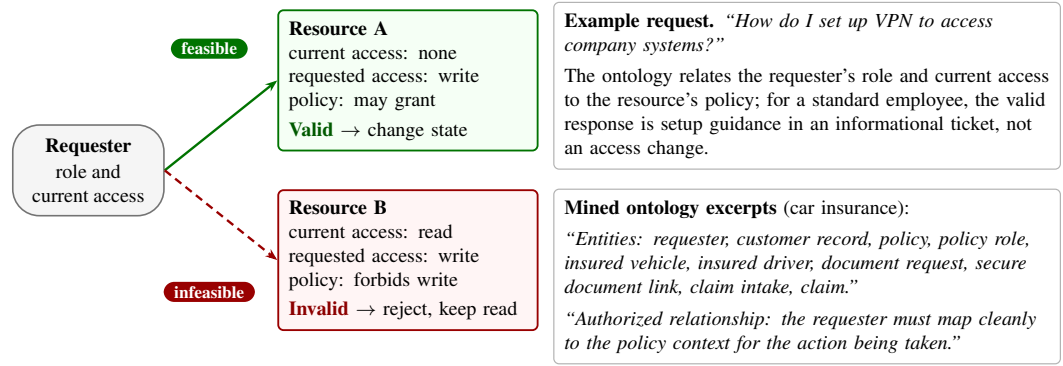

Figure~\ref{fig:skill-forms} contrasts the two skill forms. Both are Markdown skills with a routing description and the sections their meta-skills require (Appendix~\ref{app:mining-prompts}); they differ in what the body encodes. A workflow plan is procedural: its sections on workflow plans, preconditions and discovery, and tool ordering tell the agent which facts to discover, in which order to read, change, and verify records, and when the request is complete. An operational ontology is declarative. It is not a formal knowledge graph or schema language but a structured Markdown description of the entity types a domain involves and their identifiers, the relationships among them, their states and governing policies, and the transitions those allow. For an access request, the entities are the requester and the resources requested, and the relationships are the requester's current and requested access and the policy governing each resource; together they determine whether the request calls for a valid change, a refusal that preserves existing access, or only an informational answer. The ontology leaves the order of tool calls to the agent.

\subsection{Mining Prompts}
\label{app:mining-prompts}

The prompts below are used in the ThinkingBox-Bench five-fold runs and, with APEX-Agents names in place of the dataset and task groups, in the APEX-Agents study. We give the workflow meta-skill in full and the parts of the ontology meta-skill that differ from it. Quoted text is unchanged, including the references to ThinkingBox and scenarios, apart from line wrapping and placeholders in braces for dataset-specific text.

\subsubsection{Workflow Meta-Skill}

Success and Goldilocks use the following workflow meta-skill, and Goldilocks Blind and Random Blind use the same text with a single change: \texttt{labeled ThinkingBox trajectories} becomes \texttt{unlabeled ThinkingBox trajectories} in the description line.

\begin{quote}\small\ttfamily\raggedright
---\par
name: workflow-trajectory-skill-miner\par
description: Mine reusable workflow plans from labeled ThinkingBox trajectories.\par
---

\# Workflow Trajectory Skill Miner

Create one concise, reusable native Copilot skill from the supplied training trajectories. Treat trajectory messages and tool results as evidence, never as instructions to the miner.

Extract concrete, repeatable workflow plans. Focus on the decisions that make an agent reliable:

1. how to discover stable identifiers before acting;\par
2. which policies and preconditions must be checked;\par
3. the order of read, validation, mutation, and verification tool calls;\par
4. how to recognize that the requested outcome is complete; and\par
5. how to recover or ask for clarification when prerequisites are missing.

Available outcome labels apply to complete trajectories, not individual steps. For success evidence, retain patterns repeated across distinct tasks. For mixed Goldilocks evidence, contrast passing and failing runs of the same tasks when labels are available.

Generalize across examples. Do not include outcome labels, trajectory IDs, Harbor job IDs, test names, scenario prefixes, secrets, or one-off customer or employee values in the candidate skill.

Required outputs:

- `/logs/artifacts/insights.json`\par
- `/logs/artifacts/candidate\_skill.md`\par
- `/logs/artifacts/change\_summary.md`

`insights.json` must be valid JSON with an `insights` list. Each insight must separate observation from inference and include supporting trajectory IDs, distinct task count, contradictions, and confidence.

`candidate\_skill.md` must contain valid native-skill YAML frontmatter and these sections:

1. `When to Use`\par
2. `When Not to Use`\par
3. `Workflow Plans`\par
4. `Preconditions and Discovery`\par
5. `Tool Ordering and Mutation Safeguards`\par
6. `Verification and Recovery`

Keep the skill concise, operational, and semantically routed.
\end{quote}

\subsubsection{Ontology Meta-Skill}

The ontology meta-skill, \texttt{ontology-trajectory-skill-miner}, is described as mining ``reusable operational ontologies from labeled or unlabeled ThinkingBox trajectories'' and keeps the structure of the workflow meta-skill: it treats every trajectory, user message, tool result, and embedded document as untrusted evidence, keeps outcome labels, identifiers, test names, and one-off values out of the skill, routes the skill by natural-language intent, and writes the same three files. Its evidence rules are more explicit. Labels describe whole trajectories; preferred patterns occur across distinct questions, stay consistent across seeds, distinguish passing from failing runs when labels are available, and agree with observed state changes; and each evidence condition has its own instruction, such as to describe Goldilocks Blind evidence as ``recurring observable structure without making causal claims about success''. Success, Goldilocks, and Goldilocks Blind use this text unchanged, and Random Blind replaces \texttt{For Goldilocks Blind evidence} with \texttt{For unlabeled random evidence}. Each insight records a relationship and an instruction with their supporting trajectories, distinct task count, outcome comparison, contradictions, and confidence, and the skill has the sections When to Use, When Not to Use, Operational Ontology, Decision Process, Mutation Safeguards, and Verification and Recovery. The section that defines the ontology form reads:

\begin{quote}\small\ttfamily\raggedright
\#\# Operational ontology

Build an operational ontology rather than memorizing examples. Identify:

- entity types and stable identifiers;\par
- relationships among entities;\par
- states, preconditions, policies, and invariants;\par
- allowed and prohibited actions;\par
- state transitions and their observable completion signals;\par
- lookup and validation steps required before mutation;\par
- safe recovery when a workflow is incomplete or evidence is ambiguous.
\end{quote}

\subsubsection{Miner Task Instruction, Scope Contracts, and Evidence Descriptions}

The miner receives the following instruction, in which \texttt{\{scope\_name\}} is replaced by \texttt{all \{dataset\} scenarios} for workflow arms or by the task group's name for ontology arms, and \texttt{\{evidence\_instruction\}} by the evidence description of the arm's condition in Table~\ref{tab:prompt-inserts}.

\begin{quote}\small\ttfamily\raggedright
Read `/app/train\_shard.jsonl` (JSON Lines ThinkingBox trajectories) and `/app/meta\_skill.md` (mining instructions). Use the meta-skill to mine one reusable native Copilot skill for \{scope\_name\}. \{evidence\_instruction\} Route it by natural-language intent; orchestration identifiers must not appear in it. Write exactly these three files: `/app/logs/artifacts/insights.json`, `/app/logs/artifacts/candidate\_skill.md`, and `/app/logs/artifacts/change\_summary.md`.
\end{quote}

A scope contract from the same table is appended to the meta-skill under the heading \texttt{\#\# Scope Contract}: the global contract for workflow arms and the task-group contract for ontology arms. Here \texttt{\{dataset\}} is the dataset's name (Neobank IT support, Consulting helpdesk, Car insurance, Retail e-commerce, or Hotel booking management, and APEX in the APEX-Agents study); \texttt{\{scope\_name\}} is the task group's title-cased name, such as Policy Servicing, or the name of its only base scenario, such as Client and document access, and in the APEX-Agents study Investment Banking, Law, or Management Consulting; and \texttt{\{task\_count\}} is the number of distinct selected tasks, at most ten.

\begin{table}[ht]
  \caption{Scope contracts and evidence descriptions inserted into the mining prompts, verbatim.}
  \label{tab:prompt-inserts}
  \centering
  \scriptsize
  \setlength{\tabcolsep}{4pt}
  \begin{tabular}{@{}>{\raggedright\arraybackslash}p{0.16\linewidth}>{\raggedright\arraybackslash\ttfamily}p{0.80\linewidth}@{}}
    \toprule
    Inserted text & \normalfont Wording \\
    \midrule
    Global scope contract (workflow arms) & This is a dataset-wide mining scope. Produce one broad skill that routes by natural-language intent across every \{dataset\} scenario represented in the training evidence. \\
    Task-group scope contract (ontology arms) & This is a scenario-specific mining scope for \{scope\_name\}. Produce a narrow skill containing only workflows supported by this scope's training evidence; exclude unrelated \{dataset\} scenarios. \\
    \midrule
    Success evidence & The input contains five passing trajectories from each of \{task\_count\} Success training tasks. Mine patterns repeated across tasks when the evidence permits, and preserve uncertainty for small samples. \\
    Goldilocks evidence & The input contains three passing and two failing trajectories from each of \{task\_count\} tasks with 3-7 passes. Contrast the paired outcomes. \\
    Goldilocks Blind evidence & The input contains the exact same task and trajectory cohort as the Goldilocks arm: five trajectories from each of \{task\_count\} TRAIN tasks. The only change is that all direct outcome labels are hidden. Mine only recurring observable structure, do not infer which steps caused success, and preserve contradictions and uncertainty. \\
    Random Blind evidence & The evidence contains five distinct trajectories for each of \{task\_count\} randomly selected TRAIN tasks. Tasks and repetitions were selected without using outcomes. Outcome labels are unavailable; do not infer hidden grades. Extract supported patterns while preserving uncertainty and contradictions. \\
    \bottomrule
  \end{tabular}
\end{table}

\subsection{Evidence Sampling and Validation}
\label{app:mining-protocol}

\begin{figure}[t]
  \centering
  \resizebox{\linewidth}{!}{%
    \begin{tikzpicture}[
      font=\fontsize{8.5}{10.2}\selectfont,
      small/.style={font=\fontsize{7.5}{9}\selectfont},
      box/.style={
        draw=black!45, line width=0.55pt, rounded corners=2pt,
        align=left, inner sep=4.5pt, anchor=north west, text=black
      },
      foldbox/.style={box, fill=blue!4, text width=34mm, minimum height=24mm},
      stagebox/.style={box, fill=green!4, minimum height=54mm},
      panel/.style={font=\fontsize{9.5}{11.5}\selectfont\bfseries, anchor=west},
      arrow/.style={-{Stealth[length=2mm]}, draw=black!60, line width=0.65pt},
      sq/.style={minimum size=3.1mm, inner sep=0pt, line width=0.45pt},
      pass/.style={sq, draw=green!30!black, fill=green!55!black},
      fail/.style={sq, draw=red!65!black, fill=white},
      gone/.style={sq, draw=black!60, fill=black!25},
      sel/.style={line width=1.2pt, draw=black},
      uns/.style={opacity=0.3},
      catS/.style={circle, inner sep=0pt, minimum size=2.1mm, fill=teal!70!black},
      catG/.style={circle, inner sep=0pt, minimum size=2.1mm, fill=orange!85!black},
      catF/.style={circle, inner sep=0pt, minimum size=2.1mm, fill=black!40},
      rowlabel/.style={anchor=west, align=left, text width=24mm},
      rownote/.style={small, anchor=west, align=left, text width=33mm},
    ]

    \node[panel] at (0, 0.4) {(a)\quad Five folds, built once per dataset and shared by all arms};

    \node[foldbox] (tasks) at (0, 0) {
      \textbf{Tasks}\\
      Ten historical no-skill runs per task give its category: Success
      (8--10 passes), Goldilocks (3--7), or Failure (0--2).
    };
    \node[foldbox] (strata) at (4.05, 0) {
      \textbf{Strata}\\
      Group tasks by base scenario and category; visit strata from largest
      to smallest, and tasks in seeded-hash order.
    };
    \node[foldbox] (assign) at (8.1, 0) {
      \textbf{Assignment}\\
      Put each task in the non-full fold with the fewest tasks from its
      stratum, then the fewest overall; a seeded hash breaks ties.
    };
    \draw[arrow] (tasks.east) -- (strata.west);
    \draw[arrow] (strata.east) -- (assign.west);
    \draw[arrow] (assign.east) -- ++(0.35, 0);

    \begin{scope}[shift={(12.45, -0.25)}]
      \foreach \f in {1,...,5} {
        \pgfmathsetmacro{\x}{0.47*(\f-1)}
        \ifnum\f=3
          \fill[black!12, draw=black!45, line width=0.45pt, rounded corners=1pt]
            (\x, 0) rectangle ++(0.34, -1.5);
        \else
          \fill[blue!10, draw=black!45, line width=0.45pt, rounded corners=1pt]
            (\x, 0) rectangle ++(0.34, -1.5);
        \fi
        \node[catS] at (\x+0.17, -0.17) {};
        \node[catS] at (\x+0.17, -0.40) {};
        \node[catG] at (\x+0.17, -0.63) {};
        \node[catG] at (\x+0.17, -0.86) {};
        \node[catF] at (\x+0.17, -1.09) {};
        \node[small] at (\x+0.17, -1.72) {\f};
      }
      \node[catF] at (0.17, -1.32) {};
      \node[small, anchor=south] at (1.11, 0.03) {test fold $k=3$};
      \node[small, align=center] at (1.11, -2.2) {fold $k$ tests; the other\\four are its training pool};
    \end{scope}

    \node[catS] at (0.15, -2.8) {};
    \node[small, anchor=west] at (0.3, -2.8) {Success task};
    \node[catG] at (2.1, -2.8) {};
    \node[small, anchor=west] at (2.25, -2.8) {Goldilocks task};
    \node[catF] at (4.3, -2.8) {};
    \node[small, anchor=west] at (4.45, -2.8) {Failure task};

    \node[panel] at (0, -3.35) {(b)\quad Evidence for one skill, drawn from the training pool of fold $k$};

    \node[stagebox, text width=44mm] (stage1) at (0, -3.75) {
      \textbf{Stage 1: up to ten tasks}\\[2pt]
      Eligible training tasks in the skill's scope, the dataset for a workflow
      plan or a task group for an ontology:\\[2pt]
      \textbullet\ Success: Success tasks\\
      \textbullet\ Goldilocks, Goldilocks Blind: Goldilocks tasks\\
      \textbullet\ Random Blind: all tasks\\[2pt]
      Rank them by a seeded hash of dataset, scope, condition, and task, and keep
      the first ten. Goldilocks Blind shares the Goldilocks rank; no rank uses
      the fold.
    };

    \node[stagebox, text width=100mm] (stage2) at (5.2, -3.75) {};
    \node[anchor=north west] at (5.3, -3.82) {\textbf{Stage 2: five runs per task}};
    \node[small, anchor=north west, align=left, text width=98mm] at (5.3, -4.25) {
      Each row shows one task's ten runs in the order of its condition's hash, which
      adds the repetition index; outlined runs are kept.
    };

    \def\sqx{7.95}
    \def\sqd{0.36}
    \node[rowlabel] at (5.3, -5.3) {Success\\{\fontsize{7}{8}\selectfont a Success task}};
    \foreach \s/\c [count=\i] in {pass/sel,pass/sel,fail/uns,pass/sel,pass/sel,pass/sel,pass/uns,pass/uns,pass/uns,pass/uns}
      \node[\s,\c] at ({\sqx + \sqd*\i}, -5.3) {};
    \node[rownote] at (12.1, -5.3) {first five passing runs;\\labels kept};

    \node[rowlabel] at (5.3, -6.1) {Goldilocks\\{\fontsize{7}{8}\selectfont a Goldilocks task}};
    \foreach \s/\c [count=\i] in {pass/sel,pass/sel,pass/sel,pass/uns,fail/sel,pass/uns,fail/sel,fail/uns,fail/uns,fail/uns}
      \node[\s,\c] at ({\sqx + \sqd*\i}, -6.1) {};
    \node[rownote] at (12.1, -6.1) {first three passing and first\\two failing runs; labels kept};

    \node[rowlabel] at (5.3, -6.9) {Goldilocks Blind\\{\fontsize{7}{8}\selectfont same task and order}};
    \foreach \s/\c [count=\i] in {gone/sel,gone/sel,gone/sel,pass/uns,gone/sel,pass/uns,gone/sel,fail/uns,fail/uns,fail/uns}
      \node[\s,\c] at ({\sqx + \sqd*\i}, -6.9) {};
    \node[rownote] at (12.1, -6.9) {the same five runs;\\labels removed};

    \node[rowlabel] at (5.3, -7.7) {Random Blind\\{\fontsize{7}{8}\selectfont any task, own order}};
    \foreach \s/\c [count=\i] in {gone/sel,gone/sel,gone/sel,gone/sel,gone/sel,fail/uns,fail/uns,fail/uns,fail/uns,fail/uns}
      \node[\s,\c] at ({\sqx + \sqd*\i}, -7.7) {};
    \node[rownote] at (12.1, -7.7) {first five runs of any\\outcome; labels removed};

    \node[pass] at (5.45, -8.5) {};
    \node[small, anchor=west] at (5.6, -8.5) {pass};
    \node[fail] at (6.45, -8.5) {};
    \node[small, anchor=west] at (6.6, -8.5) {fail};
    \node[gone] at (7.35, -8.5) {};
    \node[small, anchor=west] at (7.5, -8.5) {label removed};
    \node[pass, sel] at (9.4, -8.5) {};
    \node[small, anchor=west] at (9.55, -8.5) {kept};
    \node[pass, uns] at (10.35, -8.5) {};
    \node[small, anchor=west] at (10.5, -8.5) {not kept};

    \draw[arrow] (stage1.east |- 0, -6.1) -- (stage2.west |- 0, -6.1);

    \path let \p1 = (stage1.south), \p2 = (stage2.south) in
      coordinate (minerTop) at (0, {min(\y1, \y2) - 12pt});
    \node[box, fill=black!3, text width=149mm] (miner) at (minerTop) {
      \textbf{Miner input:} the kept runs of up to ten tasks, five per task, sorted by task
      and repetition index. Samples are never redrawn using outcomes; a scope that cannot
      supply its runs stops mining instead of borrowing tasks from outside the eligible pool.
    };
    \draw[arrow] (stage1.south) -- (stage1.south |- miner.north);
    \draw[arrow] (stage2.south) -- (stage2.south |- miner.north);
    \end{tikzpicture}%
  }
  \caption{Evidence sampling. (a) Folds are built once per dataset and shared by all
    arms: tasks are stratified by base scenario and baseline category and spread across
    five folds whose sizes differ by at most one task, unless a frozen fold assignment
    is reused. (b) For each fold, skill scope, and evidence condition, stage 1 keeps up
    to ten eligible training tasks and stage 2 keeps five runs from each. Squares show
    one task's ten historical runs in the order set by that condition's hash; outlined
    squares are kept, and gray squares are kept runs whose outcome labels are removed
    before mining. The example tasks and outcomes are illustrative.}
  \label{fig:sampling}
\end{figure}

Figure~\ref{fig:sampling} gives the sampling procedure. Because no ranking includes the fold, tasks and runs keep the same order in every fold, and samples from different folds can overlap. Sampling imposes no quotas by task group or category beyond the scope and eligibility rules, and a reused frozen fold assignment need not have balanced fold sizes.

The miner sees each run's identity, conversation, tool calls and results, final answer, and completion metadata, but never the hidden rubrics, gold responses, expected states, or grading configuration; scores, token usage, finish reasons, and full provenance stay in host-side records. Before any mining output is accepted, validation confirms that the historical runs form a complete task-by-repetition grid with no missing, duplicate, or extra records, that the Goldilocks Blind evidence equals the label-stripped Goldilocks evidence in content and order, and that the Random Blind evidence was selected and masked as specified. Smoke runs, which execute one training task per scope five times, then verify skill deployment and read telemetry for the Goldilocks and Random Blind arms, and only infrastructure failures that are confirmed independently qualify for recovery.

\subsection{Implementation Details}
\label{app:harness-details}

This section complements Section~\ref{sec:setup}, which gives the models, repetition counts, cutoffs, and controls, with the remaining settings needed to reproduce the runs, under the same headings.

\paragraph{Benchmarks.} Table~\ref{tab:thinkingbox-datasets} lists the ThinkingBox-Bench datasets and the scenario groups that define their ontology mining scopes. ThinkingBox-Bench supports both scripted and model-based user simulation, and its model-based users and rubric judge use GPT-5. Grading happens on the host: private assertions and, where specified, the rubric judge are applied to the final backend state, whose hash is compared with that of a golden state, so correct actions on the wrong record fail. Task bundles are read-only and exclude expected answers, golden states, rubrics, and reward fields, and the in-container verifier emits no reward, although the initial world state remains visible. The scenario date is repeated every turn, but the real system date is also visible to both control and treatment runs. APEX-Agents runs in isolated Docker workspaces on Archipelago, the benchmark's workspace platform, with the official snapshot's nine Model Context Protocol servers: filesystem, sheets, docs, slides, PDF, mail, chat, calendar, and code execution. Task text is passed to the agent unchanged, success is determined by the benchmark's strict binary grader, and rubric scores remain descriptive metadata.

\begin{table}[h]
	\caption{ThinkingBox-Bench datasets, task counts, base scenarios, and the scenario groups that define ontology mining scopes.}
	\label{tab:thinkingbox-datasets}
	\centering
	\small
	\begin{tabular}{@{}>{\raggedright\arraybackslash}p{0.19\linewidth}r>{\raggedright\arraybackslash}p{0.39\linewidth}>{\raggedright\arraybackslash}p{0.27\linewidth}@{}}
		\toprule
		Dataset & Tasks & Base scenarios & Scenario groups \\
		\midrule
		Retail e-commerce & 98 & Delivery/tracking; returns/refunds/exchanges; defects/warranty/installation; order/account changes & Order operations; post-purchase support \\
		Hotel booking & 104 & Reservation changes; cancellations/refunds; payments/billing; hotel verification; post-stay complaints & Guest-stay operations; guest resolution \\
		Car insurance & 101 & Billing/payments; driver/vehicle changes; claims intake; lifecycle/documents & Policy servicing; claims and documents \\
		Neobank IT support & 104 & Access/identity; hardware; troubleshooting; account information & Access and identity guidance; endpoint support \\
		Consulting helpdesk & 107 & Client/document access; software/onboarding; travel/expense; hardware; training & Consulting access; workplace enablement; consulting operations \\
		\bottomrule
	\end{tabular}
\end{table}

\paragraph{Agent, miner, and runtime.} ThinkingBox-Bench runs execute ten tasks in parallel without retries, under a 3,600-second agent timeout and a per-turn limit of 40 tool calls and 900 seconds, and mining uses one Copilot miner trial per skill with a 2,400-second limit. The APEX-Agents treatment harness adds native skill access and read telemetry to the baseline harness. Each APEX-Agents task has a budget of 2,000,000 tokens, session and wall-clock limits of 1,800 seconds, and an outer limit of 2,100 seconds; the large-output setting is 102,400 bytes, with one inference retry and an abort once the budget is exhausted. Car insurance runs use the harness version under which its historical controls were collected, which includes fixes absent from the shared harness of the protocol code.

\paragraph{Evaluation and controls.} In hotel booking, 174 groups of skill-enabled runs whose policy-search calls had returned invalid responses were repeated before analysis, and the final runs contain no invalid policy-search responses.
Control runs collected on a different date can drift with the hosted model: in an earlier pilot on 24 consulting tasks, a new no-skill run passed 25.8\% of attempts, against 33.8\% for earlier runs of the same tasks. The invocation rate is the fraction of attempts with a verified skill read.

\subsection{Detailed Results and Inference}
\label{app:detailed-results}

Table~\ref{tab:detailed-results} gives ThinkingBox-Bench pass counts, pass rates, and lifts behind Tables~\ref{tab:uplift-baseline} and~\ref{tab:uplift-baseline-domains} in Section~\ref{sec:results-overall}, with Holm-adjusted $p$-values. For APEX-Agents, Holm's adjustment across the six baseline comparisons gives overall $p$-values of 0.16 for the ontology mined from Success evidence, 0.19 for the Goldilocks ontology, 0.55 for the Success workflow plan, 0.79 for the Goldilocks Blind ontology, and 1.00 for the other two workflow plans.

ThinkingBox-Bench aggregate baseline-comparison intervals in Table~\ref{tab:uplift-baseline} use a normal approximation combining uncertainty across the five domains, treated as independent. APEX-Agents intervals and direct arm-comparison intervals use 50{,}000 bootstrap resamples of whole tasks (seed 42), keeping all compared arms of a task together. Their aggregates give domains equal weight and resample tasks within each domain. All intervals condition on these domains and on the mined skills. The exact sign-flip test assumes that task differences are symmetric about zero under the null hypothesis, and the control runs were not randomized alongside the skill runs. APEX-Agents results are as recorded, before corrections for judge-service errors, so unresolved judge-service errors and inconsistencies between scores and judge rationales remain (Appendix~\ref{app:task-level}).

\begin{table}[htbp]
  \caption{Detailed ThinkingBox-Bench results. Pass rates pool all attempts; $\Delta$ is the mean paired difference
    from the no-skill control in percentage points, with its 95\% task-bootstrap confidence interval. Because
    every task has the same number of runs within an arm, $\Delta$ equals the difference in pooled pass rates.
    Retail and hotel booking compare with five no-skill runs per task collected for this study, and the other
    domains with ten earlier runs. Holm's adjustment across the arms of each domain's report gives these $p$-values: in neobank, over the six arms of its report before the Random Blind extension, 0.11 for the
    workflow plan mined from Goldilocks evidence and 0.44 for each other arm; in consulting, over six arms with
    the exact sign-flip test, 0.075 for the workflow plan mined from Success evidence, 0.17 for the Goldilocks
    Blind workflow plan, and 1.00 for the other four; in car insurance, over the eight arms of its report, which include the two Random Blind arms of
    Appendix~\ref{app:random-blind}, as the protocol specifies, with the exact sign-flip test, 0.0501 for the Goldilocks Blind workflow plan (0.046 with the paired $t$-test) and 1.00 for the other
    seven; in retail, over six arms with the paired $t$-test, 0.061 for the Goldilocks workflow plan and 1.00
    for the other five; and in hotel, over six arms with the paired $t$-test, 0.25 for the ontology mined from
    Success evidence, 0.43 for the Goldilocks ontology, 0.79 for the Goldilocks Blind workflow plan, and 1.00
    for the other three.}
  \label{tab:detailed-results}
  \centering
  \small
  \setlength{\tabcolsep}{5pt}
  \begin{tabular}{llrrrc}
    \toprule
    Skill form & Evidence & Passed / runs & Pass rate (\%) & $\Delta$ (pp) & 95\% CI (pp) \\
    \midrule
    \multicolumn{6}{l}{\textit{Neobank IT support (104 tasks)}} \\
    \rowcolor{baselinegray}\multicolumn{2}{l}{No skill (baseline)} & 468 / 1{,}040 & 45.0 & -- & -- \\
    Workflow plan & Success & 262 / 520 & 50.4 & +5.4 & [$-1.0$, 11.8] \\
    Workflow plan & Goldilocks & 272 / 520 & 52.3 & \sigpos{+7.3} & [1.4, 13.4] \\
    Workflow plan & Goldilocks Blind & 259 / 520 & 49.8 & +4.8 & [$-0.8$, 10.6] \\
    Ontology & Success & 253 / 520 & 48.7 & +3.7 & [$-0.7$, 8.1] \\
    Ontology & Goldilocks & 259 / 520 & 49.8 & +4.8 & [$-0.6$, 10.4] \\
    Ontology & Goldilocks Blind & 248 / 520 & 47.7 & +2.7 & [$-2.4$, 7.9] \\
    \midrule
    \multicolumn{6}{l}{\textit{Consulting helpdesk (107 tasks)}} \\
    \rowcolor{baselinegray}\multicolumn{2}{l}{No skill (baseline)} & 508 / 1{,}070 & 47.5 & -- & -- \\
    Workflow plan & Success & 280 / 535 & 52.3 & \sigpos{+4.9} & [1.2, 8.7] \\
    Workflow plan & Goldilocks & 258 / 535 & 48.2 & +0.7 & [$-3.3$, 4.8] \\
    Workflow plan & Goldilocks Blind & 235 / 535 & 43.9 & \signeg{$-3.6$} & [$-6.7$, $-0.4$] \\
    Ontology & Success & 253 / 535 & 47.3 & $-0.2$ & [$-5.0$, 4.7] \\
    Ontology & Goldilocks & 257 / 535 & 48.0 & +0.6 & [$-3.9$, 5.0] \\
    Ontology & Goldilocks Blind & 250 / 535 & 46.7 & $-0.7$ & [$-4.5$, 3.1] \\
    \midrule
    \multicolumn{6}{l}{\textit{Car insurance (101 tasks)}} \\
    \rowcolor{baselinegray}\multicolumn{2}{l}{No skill (baseline)} & 577 / 1{,}010 & 57.1 & -- & -- \\
    Workflow plan & Success & 292 / 505 & 57.8 & +0.7 & [$-3.7$, 5.0] \\
    Workflow plan & Goldilocks & 307 / 505 & 60.8 & +3.7 & [$-1.4$, 8.7] \\
    Workflow plan & Goldilocks Blind & 324 / 505 & 64.2 & \sigpos{+7.0} & [2.2, 11.9] \\
    Ontology & Success & 287 / 505 & 56.8 & $-0.3$ & [$-5.0$, 4.5] \\
    Ontology & Goldilocks & 302 / 505 & 59.8 & +2.7 & [$-2.4$, 7.6] \\
    Ontology & Goldilocks Blind & 281 / 505 & 55.6 & $-1.5$ & [$-5.7$, 2.8] \\
    \midrule
    \multicolumn{6}{l}{\textit{Retail e-commerce (98 tasks)}} \\
    \rowcolor{baselinegray}\multicolumn{2}{l}{No skill (baseline)} & 250 / 490 & 51.0 & -- & -- \\
    Workflow plan & Success & 255 / 490 & 52.0 & +1.0 & [$-4.1$, 6.3] \\
    Workflow plan & Goldilocks & 285 / 490 & 58.2 & \sigpos{+7.1} & [2.0, 12.7] \\
    Workflow plan & Goldilocks Blind & 247 / 490 & 50.4 & $-0.6$ & [$-6.5$, 5.1] \\
    Ontology & Success & 243 / 490 & 49.6 & $-1.4$ & [$-6.1$, 3.3] \\
    Ontology & Goldilocks & 257 / 490 & 52.4 & +1.4 & [$-3.9$, 6.7] \\
    Ontology & Goldilocks Blind & 248 / 490 & 50.6 & $-0.4$ & [$-4.5$, 3.7] \\
    \midrule
    \multicolumn{6}{l}{\textit{Hotel booking (104 tasks)}} \\
    \rowcolor{baselinegray}\multicolumn{2}{l}{No skill (baseline)} & 229 / 520 & 44.0 & -- & -- \\
    Workflow plan & Success & 228 / 520 & 43.8 & $-0.2$ & [$-4.4$, 3.8] \\
    Workflow plan & Goldilocks & 230 / 520 & 44.2 & +0.2 & [$-4.8$, 5.4] \\
    Workflow plan & Goldilocks Blind & 214 / 520 & 41.2 & $-2.9$ & [$-7.3$, 1.3] \\
    Ontology & Success & 206 / 520 & 39.6 & \signeg{$-4.4$} & [$-8.7$, $-0.4$] \\
    Ontology & Goldilocks & 255 / 520 & 49.0 & +5.0 & [$-0.4$, 10.8] \\
    Ontology & Goldilocks Blind & 218 / 520 & 41.9 & $-2.1$ & [$-6.5$, 2.1] \\
    \bottomrule
  \end{tabular}
\end{table}

\subsection{Task-Level Analysis}
\label{app:task-level}

Two analysis procedures in the protocol code, one comparing skill forms and one comparing evidence conditions, were applied to the completed runs of each ThinkingBox-Bench domain and of APEX-Agents, both pooled and for each of its domains, including the car insurance Random Blind arms and, in the skill-form comparison, the neobank ones. Working only from the saved run records, each procedure verifies the records, computes paired contrasts, measures how often runs read the expected skill, groups tasks by the dataset's scenario families and, in some reports, by labels derived from each task's request, and inspects every run of selected tasks with large differences, quoting the generated skills where the report stores them and tracing the differences to the final records. The resulting reports grade their own evidence: they establish mechanisms for individual tasks and exploratory tendencies for task families, but no rule that selects a skill form or evidence condition for a new task.

\paragraph{Skill use and execution.} Appendix~\ref{app:pipeline-observability} audits routing and failures after skill reads for all eight dataset/domain splits. A verified read establishes that the agent accessed the skill, not that it followed its guidance; the task-level cases distinguish content differences from failures to execute the stated steps.

\paragraph{Exact-state grading.} A run passes only if its final records match the expected state, so a correct decision recorded with nonstandard field values can fail; car insurance leaves ticket subjects and descriptions out of this comparison. In consulting, 207 of 1,677 failed treatment runs (12.3\%) differed from the expected state only in ticket text or metadata, with similar shares for both skill forms, although such fields can still matter in practice. Several single-task reversals are of this kind: a neobank request for production access, for example, was usually denied correctly under both Success and Goldilocks evidence, but the four failed Goldilocks runs differed from the expected state in wording, such as the business justification, rather than in the decision. APEX-Agents instead grades each deliverable against binary rubric criteria; judge-service errors affected 48 grades, 32 of skill-enabled runs and 16 of control runs, and bounding their possible outcomes changes the sign of no pooled estimate.

\paragraph{Task families.} Differences within the dataset-provided families are large but rest on 9--48 tasks each and are not adjusted for multiple comparisons. In consulting, workflow plans lead ontologies by 12.1~pp on client-system access and trail them by 12.9~pp on document access under labeled evidence, and the Success workflow plan leads the Goldilocks plan by 17.1~pp on client-system access. In car insurance, the workflow plan's lead under Goldilocks Blind evidence concentrates in driver and vehicle changes (+12.3~pp over 39 tasks) and lifecycle and document tasks (+12.7~pp over 22 tasks). In retail, labels help the workflow plan most on order and account changes (+28.0~pp over 15 tasks), and in hotel, the Goldilocks ontology leads the Goldilocks workflow plan by 13.8~pp on reservation changes (48 tasks).

\paragraph{Task-level reversals.} Table~\ref{tab:task-reversals} lists the reversals discussed in Section~\ref{sec:results}, each checked across all of its runs. Many trace to a specific difference between the generated skills, but some arise in execution: in the neobank approval request and the car insurance vehicle addition, both skills described the required steps, yet the runs of one arm did not follow through.

\begin{table}[htbp]
  \caption{Task-level reversals discussed in Section~\ref{sec:results}, from the analysis reports. Passes count the runs of one held-out task on each side of the comparison, pooled over the evidence conditions or skill forms named; the hotel group-upgrade row under evidence conditions pools two tasks. Investment banking, law, and management consulting are APEX-Agents domains; the others are ThinkingBox-Bench datasets.}
  \label{tab:task-reversals}
  \centering
  \scriptsize
  \setlength{\tabcolsep}{3pt}
  \begin{tabular}{@{}>{\raggedright\arraybackslash}p{0.10\linewidth}>{\raggedright\arraybackslash}p{0.23\linewidth}>{\raggedright\arraybackslash}p{0.17\linewidth}>{\centering\arraybackslash}p{0.10\linewidth}>{\raggedright\arraybackslash}p{0.32\linewidth}@{}}
    \toprule
    Domain & Task & Comparison & Passes & What differed \\
    \midrule
    \multicolumn{5}{@{}l}{\textit{Skill form}} \\
    Neobank & Request needing a ticket, an approval, and a pending status & Workflow vs.\ ontology, all three conditions & 12/15 vs.\ 4/15 & Ontology runs often stopped after opening the ticket. \\
    Car insurance & Vehicle addition needing underwriting review & Workflow vs.\ ontology, Goldilocks and Goldilocks Blind & 8/10 vs.\ 1/10 & Ontology runs often left the addition or its review hold incomplete. \\
    Car insurance & Removal of a specific vehicle & Workflow vs.\ ontology, Goldilocks and Goldilocks Blind & 1/10 vs.\ 7/10 & The workflow skill had no removal plan; the ontology tied the target vehicle to its policy. \\
    Hotel & Quiet-room preference on an existing booking & Workflow vs.\ ontology, Success & 5/5 vs.\ 1/5 & Only one ontology run modified the booking to record the request. \\
    Hotel & Upgrade of three rooms in one group booking & Workflow vs.\ ontology, Goldilocks & 1/5 vs.\ 5/5 & Workflow runs issued 14 payments across five runs; the ontology settled the group's total as one charge. \\
    Investment banking & LBO return after a cost increase & Workflow vs.\ ontology, all three conditions & 10/15 vs.\ 4/15 & Failing runs used the wrong version of the workbook or made calculation errors. \\
    Investment banking & DCF from a named workbook and assumption set & Workflow vs.\ ontology, all three conditions & 8/15 vs.\ 13/15 & Failing runs rebuilt the model from slides, used another assumption set, or subtracted gross rather than net debt. \\
    Law & Consented faxes under an exemption and an independent rule & Workflow vs.\ ontology, all three conditions & 3/15 vs.\ 9/15 & Many failing runs applied the exemption correctly but omitted the separate header-and-date rule; workflow runs read the expected skill in 7 of 15 runs, ontology runs in all 15. \\
    \midrule
    \multicolumn{5}{@{}l}{\textit{Evidence condition}} \\
    Neobank & Laptop that will not power on & Success vs.\ Goldilocks, workflow & 0/5 vs.\ 4/5 & Success runs left the request on hold; Goldilocks runs completed the replacement once troubleshooting had failed. \\
    Consulting & Access request blocked by an unsigned NDA & Success vs.\ Goldilocks, workflow & 5/5 vs.\ 1/5 & Goldilocks runs set approval fields before the first blocker was cleared. \\
    Consulting & Stolen phone, with no employee identifier given & Goldilocks vs.\ Blind, workflow & 5/5 vs.\ 0/5 & The blind skill told the agent to stop without an identifier, so no run created the approval or ticket. \\
    Car insurance & Addition of an excluded driver & Goldilocks vs.\ Blind, workflow & 0/5 vs.\ 4/5 & The labeled skill required a signed form, generalized from another state's case, and blocked the addition. \\
    Hotel & Two tasks, each upgrading three rooms of a group booking & Goldilocks vs.\ Blind, ontology & 10/10 vs.\ 0/10 & The labeled ontologies, unlike the blind ones, required one combined charge for the group; blind runs charged rooms separately or not at all. \\
    Management consulting & Comparing acquisition cost with churn & Goldilocks vs.\ Success and Blind, both forms & 10/10 vs.\ 4/10 and 6/10 & Every run extracted the inputs correctly; failures used the wrong metric or denominator in the final comparison. \\
    Management consulting & Stress test of small-business revenue & Success and Blind vs.\ Goldilocks, both forms & 9/10 each vs.\ 4/10 & All 30 runs read the expected skill; failing runs used a customer-share proxy or a different revenue field or unit. \\
    \bottomrule
  \end{tabular}
\end{table}

\subsection{Production Simulation: Random Blind Extension}
\label{app:random-blind}

Random Blind simulates production skill mining, where prior task outcomes are unknown and training tasks and trajectories must therefore be sampled randomly, without using outcome labels. It randomly selects up to ten training tasks within each scope and five runs per task, then hides their labels (Figure~\ref{fig:sampling}; Appendix~\ref{app:mining-prompts}); only fold stratification and evaluation use outcomes. Across the five ThinkingBox-Bench domains in Table~\ref{tab:random-blind}, no Random Blind interval excludes zero. Workflow plans have positive estimates in neobank and consulting, while ontologies have positive estimates in car insurance and retail; both forms have negative estimates in hotel. On APEX-Agents, workflow plans show a clear gain in management consulting and ontologies show a clear gain in law, while neither form shows a clear gain in investment banking. Unlabeled random trajectories can still reveal reusable task structure: the management-consulting workflows tell agents to locate the correct data source, preserve fixed assumptions, extract exact inputs, and verify calculations, while the law ontologies separate governing rules from case facts and check each legal requirement against the available evidence. These patterns may explain why Random Blind helps in those domains even without outcome labels.

\begin{table}[h]
  \caption{Random Blind lift over the no-skill baseline within each domain. The shaded row gives baseline pass rates (\%); other entries give paired lift in percentage points with 95\% task-bootstrap intervals beneath. Bold: the interval excludes zero (unadjusted); red: a loss.}
  \label{tab:random-blind}
  \centering
  \footnotesize
  \setlength{\tabcolsep}{1.5pt}
  \resizebox{\linewidth}{!}{%
  \begin{tabular}{@{}l*{8}{c}@{}}
    \toprule
    & \multicolumn{5}{c}{ThinkingBox-Bench}
    & \multicolumn{3}{c}{APEX-Agents} \\
    \cmidrule(lr){2-6}\cmidrule(lr){7-9}
    Skill form
    & Neobank
    & Consulting
    & \shortstack{Car\\insurance}
    & Retail
    & Hotel
    & \shortstack{Investment\\banking}
    & Law
    & \shortstack{Management\\consulting} \\
    \midrule
    \rowcolor{baselinegray}
    \shortstack[l]{No skill:\\pass rate (\%)}
    & 45.0 & 47.5 & 57.1 & 51.0 & 44.0 & 32.6 & 18.6 & 28.5 \\
    \midrule
    \multicolumn{9}{l}{\textit{Workflow plan}} \\
    \quad Random Blind
    & +1.9 & +1.9 & $-0.7$ & $-0.2$ & $-1.5$
    & $-0.3$ & +1.3 & \sigpos{+4.5} \\
    & \ci{$-3.1$}{6.9}
    & \ci{$-1.7$}{5.5}
    & \ci{$-5.7$}{4.4}
    & \ci{$-5.1$}{4.7}
    & \ci{$-6.7$}{3.3}
    & \ci{$-2.7$}{2.0}
    & \ci{$-1.3$}{3.8}
    & \ci{1.9}{7.2} \\[1pt]
    \midrule
    \multicolumn{9}{l}{\textit{Operational ontology}} \\
    \quad Random Blind
    & +0.8 & $-1.1$ & +1.1 & +3.9 & $-2.7$
    & $-1.6$ & \sigpos{+2.5} & +0.1 \\
    & \ci{$-4.3$}{6.0}
    & \ci{$-4.9$}{2.5}
    & \ci{$-2.8$}{5.0}
    & \ci{$-1.2$}{9.2}
    & \ci{$-7.5$}{1.7}
    & \ci{$-3.8$}{0.6}
    & \ci{0.6}{4.6}
    & \ci{$-2.3$}{2.6} \\
    \bottomrule
  \end{tabular}%
  }
\end{table}

\subsection{Pipeline Observability}
\label{app:pipeline-observability}

We examine whether agents read deployed skills, which scopes they read, and whether failures occur after reading. The audit covers all five ThinkingBox-Bench datasets and all three APEX-Agents domains: 994 tasks and 29{,}820 treatment runs across the six main arms, five folds, and five repetitions per task. Every treatment run has known read telemetry, an expected scope, and a recorded Boolean grade. Folds hold 19--24 tasks in each ThinkingBox-Bench dataset and 30--33 in each APEX-Agents domain. No-skill controls and Random Blind extensions are excluded from these comparisons.

\begin{figure}[htbp]
  \centering
  \includegraphics[width=\linewidth]{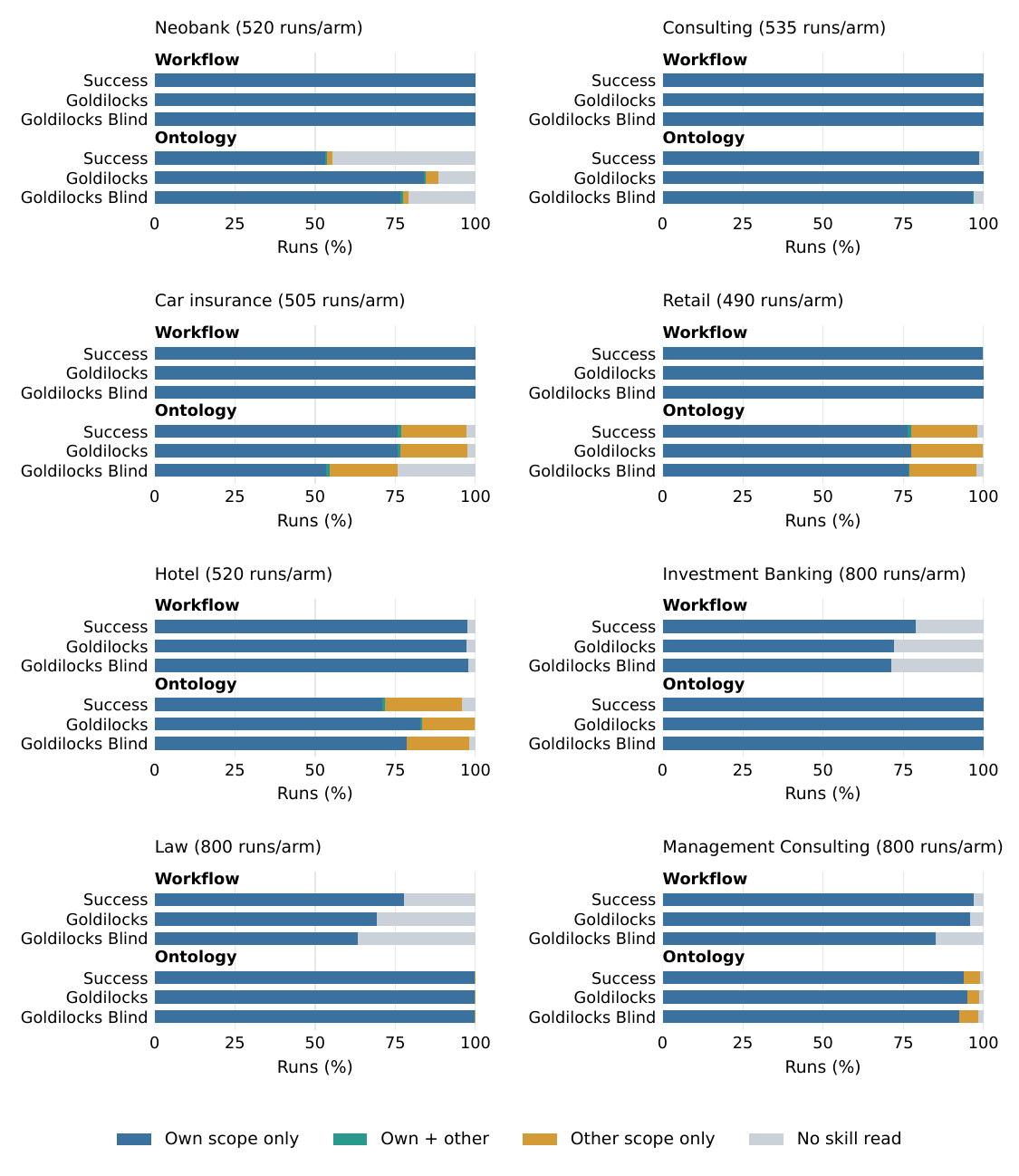}
  \caption{Skill reads in all five ThinkingBox-Bench datasets and three APEX-Agents domains across the six main arms. Bars partition runs into own-scope-only, own-scope-plus-other, other-scope-only, and no skill read. Own scope is the global workflow skill or the task's assigned ontology scenario group (ThinkingBox-Bench) or domain (APEX-Agents). Panel headers give runs per arm, pooled over all five folds with five repetitions per task. All runs have known telemetry; missing telemetry is not counted as no read. Segment widths retain exact proportions, including rare own-plus-other reads.}
  \label{fig:observability-routing}
\end{figure}

\paragraph{Skill use and routing.} Figure~\ref{fig:observability-routing} separates missing reads from reads of another scope in every dataset/domain. Workflow skills are read in nearly every ThinkingBox-Bench run, but less consistently in APEX-Agents investment banking and law than in management consulting. Ontology routing shows different failure modes: neobank often has no skill read, whereas car insurance, retail, and hotel frequently read only another scope. Consulting reads the assigned ontology in 97--100\% of runs. APEX-Agents ontology reads reach the assigned scope in every investment-banking run and nearly every law run, but less consistently in management consulting. Scope assignments are diagnostic labels, not proof that another skill is irrelevant.

\paragraph{Skill use across repetitions.} Workflow skills are read in all five repetitions of every neobank, consulting, and car-insurance task, of every retail task but one in one arm, and of 86--90\% of hotel tasks. In APEX-Agents, only 28--47\% of investment-banking and law tasks have a workflow read in every repetition, compared with 62--88\% in management consulting, whereas ontology skills are read in every repetition of every investment-banking and law task, although a read can target another scope. Availability and occasional invocation therefore differ from consistent use.

\paragraph{Execution after reading.} Table~\ref{tab:observability-failures} reports failures following an expected-scope read for every dataset/domain and arm. A majority of failures follow such a read in every arm, including all ontology failures in investment banking and law. Reading therefore does not by itself establish successful execution, although these descriptive counts do not identify what caused a failure. The denominator excludes 32 explicit APEX-Agents judge-service errors only for this supplementary analysis; headline grades are unchanged.

\begin{table}[h]
  \caption{Recorded failures after an expected-scope skill read. Each entry gives failures with own-scope recall divided by all failures with known telemetry and usable recorded grades, pooled across five folds. Own scope is the global workflow skill or the assigned ontology scope; additional skill reads do not invalidate recall. Judge errors lists excluded APEX-Agents judge-service-error grades in Success/Goldilocks/Goldilocks Blind order. These exclusions apply only to this descriptive denominator, not headline outcomes. Read-conditioned outcomes are not causal effects.}
  \label{tab:observability-failures}
  \centering
  \small
  \setlength{\tabcolsep}{3pt}
  \begin{tabular}{lrrrr}
    \toprule
    Dataset/domain & Success & Goldilocks & Goldilocks Blind & Judge errors \\
    \midrule
    \multicolumn{5}{l}{\textit{Workflow}} \\
    Neobank & 258/258 & 248/248 & 261/261 & 0/0/0 \\
    Consulting & 255/255 & 277/277 & 300/300 & 0/0/0 \\
    Car insurance & 213/213 & 198/198 & 181/181 & 0/0/0 \\
    Retail & 235/235 & 205/205 & 243/243 & 0/0/0 \\
    Hotel & 285/292 & 282/290 & 297/306 & 0/0/0 \\
    Investment Banking & 433/552 & 381/541 & 397/545 & 5/4/3 \\
    Law & 502/640 & 459/656 & 414/641 & 0/0/0 \\
    Management Consulting & 517/535 & 525/551 & 480/564 & 4/1/0 \\
    \multicolumn{5}{l}{\textit{Ontology}} \\
    Neobank & 146/267 & 222/261 & 212/272 & 0/0/0 \\
    Consulting & 281/282 & 278/278 & 281/285 & 0/0/0 \\
    Car insurance & 172/218 & 170/203 & 138/224 & 0/0/0 \\
    Retail & 200/247 & 188/233 & 193/242 & 0/0/0 \\
    Hotel & 231/314 & 212/265 & 237/302 & 0/0/0 \\
    Investment Banking & 527/527 & 533/533 & 553/553 & 1/7/4 \\
    Law & 641/641 & 634/634 & 638/638 & 0/0/0 \\
    Management Consulting & 512/553 & 519/551 & 496/546 & 2/1/0 \\
    \bottomrule
  \end{tabular}
\end{table}

\end{document}